\documentclass{article}
\usepackage[utf8]{inputenc}
\usepackage{amsmath,amssymb}
\usepackage{graphicx}
\usepackage{booktabs}
\usepackage{hyperref}
\usepackage{xcolor}
\usepackage[margin=1in]{geometry}
\usepackage{subcaption}
\usepackage{multirow}
\usepackage{algorithm}
\usepackage{algorithmic}

\graphicspath{{./}{figures/}}  

\title{HubRouter: A Pluggable Sub-Quadratic Routing Primitive\\for Hybrid Sequence Models}

\author{Abhinaba Basu\thanks{Corresponding author: \texttt{mail@abhinaba.com}}\\
\small National Institute of Electronics and Information Technology (NIELIT)}

\date{}

\begin{document}
\maketitle

\begin{abstract}
We introduce \textbf{HubRouter}, a pluggable module that can replace $O(n^2)$ attention layers with $O(nM)$ hub-mediated routing, where $M \ll n$ is a small number of learned hub tokens. We demonstrate it in two architectures from scratch (a Jamba-style hybrid and a 12-layer Transformer); broader architectural generality is not tested here.
HubRouter implements an encode--decode--score--council pipeline: $M$ learned hubs cross-attend to all tokens (encode), tokens project against hubs to obtain routing fingerprints (decode), a score head selects top-$k$ tokens, and a sparse council attends only to the selected subset.
In the two architectures we test, the module slots in at the attention-layer interface and is trained from scratch; retrofit into pretrained models is a tested negative case discussed as a limitation (Section~\ref{sec:discussion}).

We validate HubRouter in three settings:
(1)~\textbf{Hub-Jamba} replaces attention layers in a Jamba-style hybrid, yielding a nominal 4.2\% PPL improvement (200.2 vs~209.0, single seed; the gap may be weakly within seed noise---see Section~\ref{sec:discussion} Limitation 5) with up to $\sim$90$\times$ training throughput at sequence length~1024 in matched PyTorch-native implementations; a kernel-optimised baseline would narrow this multiplier to an estimated $\sim$10--15$\times$ (Section~\ref{sec:discussion} Limitation 6).
(2)~\textbf{Graduated replacement} of 25\% of Transformer attention layers yields the best perplexity in our matched-budget sweep (268.0 vs 282.4 pure Transformer, 3000 training steps), characterizing the quality--efficiency sweet spot in this regime.
(3)~\textbf{Hub-GPT} provides strictly causal hub routing for autoregressive LM, achieving PPL~$211.5{\pm}0.4$ (3 seeds, post council-causal fix); approximately 3 PPL worse than Jamba's $208.5{\pm}0.7$, a measurable quality cost for avoiding $O(n^2)$ computation. Post-fix, chunk size $C$ has little effect (all $C$ cluster at 211.5--212.8 PPL); the pre-fix chunk-size benefit was an artifact of a bidirectional-council leak we found in adversarial review and disclose in Section~\ref{sec:discussion}.

A multi-seed routing-task sweep (13 no-ortho cells and 8 ortho cells across $M{=}1$--$32$, 5 seeds each, $\sim$105 runs total) reveals that $M{=}8$--$14$ reliably converges ($4$--$5$/$5$ seeds without ortho), with $M{=}6$ rescued to $5$/$5$ by orthogonal regularization; $M{=}16$ is the standard choice from prior work~\cite{basu2024routing} and is cited but not re-run here, while $M{\geq}20$ shows increasing seed sensitivity.
Orthogonal regularization cleanly rescues $M{=}6$ (4/5$\rightarrow$5/5). At $M{\geq}8$ its effect is inconsistent: marginally harmful at $M{=}12$ (4/5$\rightarrow$3/5, two cells drop just below the 90\% threshold), neutral in success count at $M{=}14$ and $M{=}20$ (though $M{=}20$'s mean routing drops from 91.8\% to 78.2\%), worse at $M{=}24$ (3/5$\rightarrow$2/5), and partially helpful at $M{=}32$ (3/5$\rightarrow$4/5).
All code and experiment scripts will be made publicly available with the final version.
\end{abstract}

\section{Introduction}

Hybrid sequence models --- combining cheap recurrence (Mamba~\cite{gu2023mamba}, RWKV~\cite{peng2023rwkv}) with selective attention --- have emerged as a leading paradigm for efficient long-context modeling~\cite{lieber2024jamba,de2024griffin,glorioso2024zamba}.
The key design decision is \emph{which tokens receive expensive attention}.
This routing decision is typically implicit: fixed schedules interleave recurrent and attention layers at predetermined intervals (every 5th layer in Jamba, every 6th in Griffin).

We argue that routing should be \emph{learned and content-based} rather than fixed. Concretely: if a sequence contains a rare associative-recall query (e.g., a tool-output identifier whose value is referenced much later), fixed-interval schedules will waste attention on layers where it is not needed and miss it where it is. Content-based routing adapts to the sequence.
Prior work~\cite{basu2024routing} established that content-based routing requires pairwise token comparison --- an apparently $O(n^2)$ operation that seems to defeat the purpose of routing.
However, we show that \textbf{hub-mediated routing} can serve as a cheaper surrogate, running in $O(nM)$ time by compressing token information through $M$ learned hub vectors before making routing decisions. This is not a proof of equivalence to pairwise comparison; we show empirically that the cheaper surrogate suffices in the settings tested.

\paragraph{Contributions.}
\begin{enumerate}
    \item \textbf{HubRouter module}: A self-contained module with $O(nM + k^2 d)$ complexity (Section~\ref{sec:method}). Drop-in for from-scratch training; retrofit limitations discussed in Section~\ref{sec:discussion}.
    \item \textbf{Hub-Jamba}: Replacing attention in Jamba-style hybrids yields a 4.2\% PPL gain (single seed) and 25$\times$--183$\times$ training throughput over matched PyTorch-native baselines (Section~\ref{sec:hubjamba}; kernel caveats in Section~\ref{sec:discussion}).
    \item \textbf{Hub-GPT}: Chunked causal encoding plus a causally-masked council enables autoregressive LM without $O(n^2)$, at a measured $\approx$3 PPL quality cost vs attention-based Jamba (Section~\ref{sec:hubgpt}).
    \item \textbf{Graduated replacement}: 25\% replacement achieves the quality--efficiency sweet spot at matched training budget; the full curve characterizes the tradeoff (Section~\ref{sec:graduated}).
    \item \textbf{M-sweep}: Multi-seed study ($\sim$105 runs) maps reliable convergence at $M{=}8$--$14$, with orthogonal regularization rescuing $M{=}6$; $M{=}16$ is the prior-work default~\cite{basu2024routing} (Section~\ref{sec:msweep}).
\end{enumerate}

\section{HubRouter Module}
\label{sec:method}

\subsection{Architecture}

HubRouter implements a four-stage pipeline operating on input representations $X \in \mathbb{R}^{n \times d}$ (Figure~\ref{fig:architecture}):

\begin{figure}[t]
    \centering
    \includegraphics[width=\textwidth]{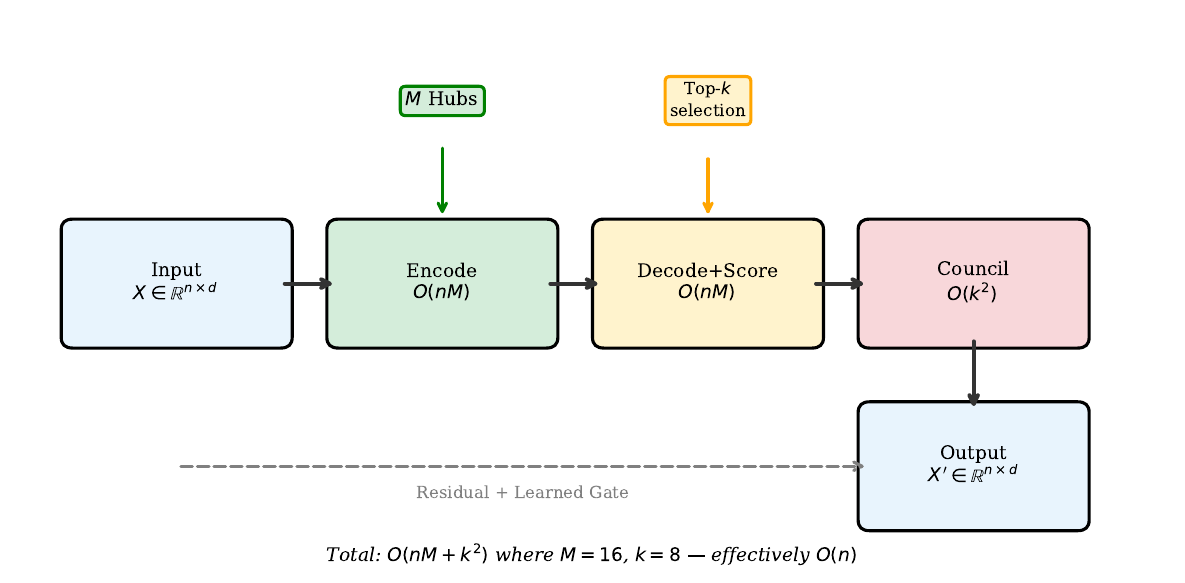}
    \caption{\textbf{HubRouter pipeline.} Input tokens are compressed through $M$ learned hubs (encode, $O(nM)$), tokens project against hubs for routing fingerprints (decode, $O(nM)$), a score head selects top-$k$ tokens, and a council applies sparse attention only to the selected subset ($O(k^2)$). A learned gate fuses council output back into the residual stream. \emph{Key insight:} routing decisions are made through $M$-dimensional hub space rather than the $n{\times}n$ pairwise comparison matrix, converting a quadratic bottleneck into a learned low-rank one.}
    \label{fig:architecture}
\end{figure}

\paragraph{Stage 1: Encode.}
$M$ learned hub embeddings $H \in \mathbb{R}^{M \times d}$ cross-attend to all $n$ tokens:
\begin{equation}
    H' = H + \text{MultiHead}(Q{=}H, K{=}X, V{=}X) \quad \in \mathbb{R}^{M \times d}
\end{equation}
Cost: $O(nMd)$. Each hub absorbs information from the entire sequence, forming a compressed global summary.

\paragraph{Stage 2: Decode.}
Each token computes a routing fingerprint by projecting against the enriched hubs:
\begin{equation}
    F_i = \text{softmax}\left(\frac{x_i \cdot H'^{\top}}{\sqrt{d}}\right) H' \quad \in \mathbb{R}^d
\end{equation}
Cost: $O(nMd)$. The fingerprint $F_i$ encodes token $i$'s relationship to all hubs.

\paragraph{Stage 3: Score and Select.}
An MLP scores each token's routing importance: $s_i = \text{MLP}(F_i) \in \mathbb{R}$.
The top $k/2$ tokens by score are selected, then expanded with their right-neighbors (selecting token $j$ also selects $j{+}1$), yielding up to $k$ tokens total:
\begin{equation}
    \mathcal{S} = \text{TopK}_{k/2}\!\left(\{s_i\}_{i=1}^n\right) \cup \left\{j{+}1 \,:\, j \in \text{TopK}_{k/2}\right\}
\end{equation}
Cost: $O(nd + n\log k)$.

\paragraph{Stage 4: Council.}
Standard multi-head self-attention restricted to the $k$ selected tokens, with a feed-forward residual:
\begin{equation}
    Y_{\mathcal{S}} = \text{MultiHead}(Q{=}X_\mathcal{S}, K{=}X_\mathcal{S}, V{=}X_\mathcal{S}) + \text{FFN}(\text{MultiHead}(\cdot))
\end{equation}
Cost: $O(k^2 d)$ where $k \ll n$ (typically $k{=}8$).

\paragraph{Fusion.}
A learned gate blends council output back into the residual stream:
$x_i' = x_i + \sigma(g) \cdot Y_i$ for $i \in \mathcal{S}$, identity for $i \notin \mathcal{S}$.

\paragraph{Total complexity: $O(nMd + k^2d)$.}
For fixed $M{=}16$, $k{=}8$, this is linear in $n$ with small constants, versus $O(n^2 d)$ for standard attention. We use $O(nM + k^2)$ as shorthand throughout when the $d$ factor is shared with the attention baseline.

\paragraph{Worked example.} To anchor the pipeline concretely, consider $n{=}8$ tokens, $M{=}2$ hubs, $k{=}4$, $d{=}64$. Stage~1 produces $H' \in \mathbb{R}^{2 \times 64}$ where each of the two hub rows is a learned weighted summary of all 8 tokens (e.g., hub $h_1$ may emphasize local-neighbour content, hub $h_2$ rare-token content). Stage~2 gives each token a fingerprint $F_i \in \mathbb{R}^{64}$ that is a softmax-weighted mix of $(h_1, h_2)$; tokens playing similar roles get similar fingerprints. Stage~3 picks the 2 highest-scored tokens, adds their right-neighbours, forming $\mathcal{S}$ with $|\mathcal{S}|{=}4$. Stage~4 runs a standard $4 \times 4$ attention on those 4 tokens. The $8{\times}8$ attention matrix is never materialised; the routing decision is made in 2-dimensional hub space.

\subsection{Chunked Causal Encoding for LM}
\label{sec:causal}

For autoregressive language modeling, bidirectional hub encoding would leak future information.
We introduce \textbf{chunked causal encoding}: the sequence is divided into chunks of size $C$, and hub states are carried forward cumulatively with a learned gate:

\begin{equation}
    H'_k = H'_{k-1} + \sigma(g_k) \cdot \text{MultiHead}(H'_{k-1}, X_{[kC:(k{+}1)C]}, X_{[kC:(k{+}1)C]})
\end{equation}

After processing chunk $k$ (0-indexed), hubs have absorbed information from chunks $0 \ldots k$ only.
Tokens in chunk $k$ decode against $H'_k$, ensuring no leakage from future chunks.
Within a chunk, the encode step is bidirectional, so a token at position $kC + j$ can be informed (through hubs) by any other token in $[kC, kC{+}C)$; this is bounded but nonzero. Setting $C{=}1$ degenerates the chunk to a single token, eliminating within-chunk leakage entirely.

\subsection{Drop-In API}

HubRouter exposes the same interface as a Transformer attention layer:
\begin{verbatim}
router = HubRouter(d_model=768, n_hubs=16, n_heads=4, top_k=8,
                   chunk_size=64)  # None for bidirectional
output = router(hidden_states)  # (B, L, D) -> (B, L, D)
\end{verbatim}

This enables substitution in most architectures that use attention layers, when training from scratch (see retrofit caveat in Section~\ref{sec:discussion}).

\section{Hub-Jamba: Replacing Attention in Hybrids}
\label{sec:hubjamba}

Jamba~\cite{lieber2024jamba} interleaves Mamba layers with attention at fixed intervals.
Hub-Jamba replaces every attention layer with a HubRouter block, keeping the Mamba layers unchanged.

\subsection{Setup}

All models are trained for 3000 steps on WikiText-103 with identical hyperparameters (AdamW, lr=$3{\times}10^{-4}$, batch size 16, sequence length 256, 500-step linear warmup, no LR decay, gradient clip 1.0). Hub-Jamba PPL in Table~\ref{tab:hubjamba} is single-seed; 3-seed robustness for the Jamba baseline on this setup is characterized separately in Section~\ref{sec:hubgpt}.
Parameter budgets are approximately matched: Jamba (17.8M), Hub-Jamba M=4 (20.0M, +12\%), Hub-Jamba M=16 (20.0M, +12\%); the parameter difference is from the additional $M$ hub vectors and is small relative to the architectural difference being tested.

\subsection{Results}

\begin{table}[h]
\centering
\caption{\textbf{Hub-Jamba vs Jamba.} Replacing attention with HubRouter improves PPL and provides large throughput gains under matched PyTorch-native baselines (kernel caveats in Section~\ref{sec:discussion}). PPL values are single-seed; the 3-seed mean for the Jamba baseline (Table~\ref{tab:chunk}) is 208.5$\pm$0.7, indicating the 4.2\% Hub-Jamba improvement is at most weakly above seed noise. SNLI is evaluated by training a small classifier on frozen LM hidden states (zero-shot 3-way SNLI is at 33.3\% chance; both models cluster near chance, so SNLI numbers should not be over-interpreted). Throughput measured at seq=1024.}
\label{tab:hubjamba}
\begin{tabular}{lccccc}
\toprule
\textbf{Model} & \textbf{PPL}$\downarrow$ & \textbf{SNLI} & \textbf{Train tok/s} & \textbf{Infer tok/s} & \textbf{Params} \\
\midrule
Jamba-style & 209.0 & 35.3\% & 320 & 2,603 & 17.8M \\
Hub-Jamba M=16 & 201.1 & 35.4\% & 27,836 & 77,797 & 20.0M \\
Hub-Jamba M=4 & \textbf{200.2} & \textbf{36.8\%} & \textbf{28,844} & 73,089 & 20.0M \\
\bottomrule
\end{tabular}
\end{table}

\begin{figure}[t]
    \centering
    \includegraphics[width=\textwidth]{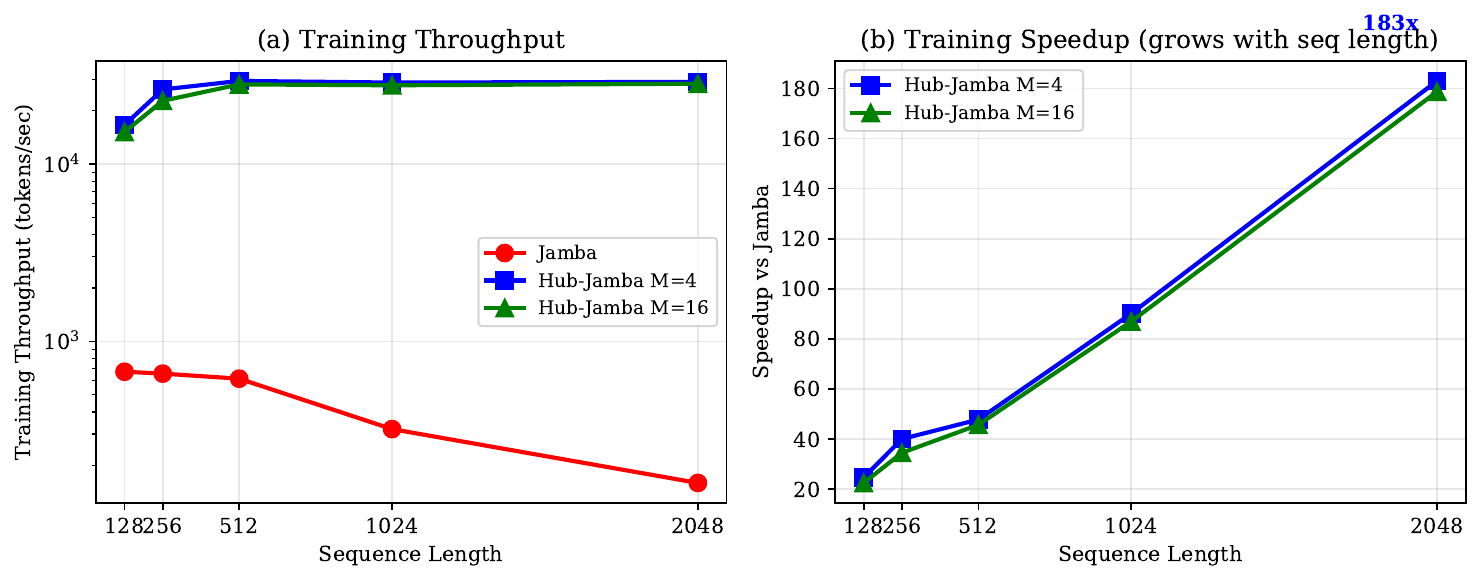}
    \caption{\textbf{Throughput scaling (matched PyTorch-native baselines).} (a)~Hub-Jamba maintains constant throughput as sequence length grows (sub-quadratic), while Jamba degrades due to quadratic attention. (b)~Speedup increases with sequence length, reaching $\sim$183$\times$ at seq=2048. Caveat (Section~\ref{sec:discussion}): the Jamba baseline does not use FlashAttention or production Mamba CUDA kernels.}
    \label{fig:throughput}
\end{figure}

Hub-Jamba achieves a 4.2\% nominal perplexity improvement (200.2 vs 209.0; single-seed Hub-Jamba vs single-seed Jamba in Table~\ref{tab:hubjamba}) with $\sim$90$\times$ training throughput at sequence length 1024 in matched PyTorch-native implementations.
Sub-quadratic throughput scaling holds across sequence lengths (Table~\ref{tab:throughput} and Figure~\ref{fig:throughput}).

\begin{table}[h]
\centering
\caption{\textbf{Throughput scaling with sequence length.} Hub-Jamba maintains constant throughput as sequence length grows (sub-quadratic), while Jamba degrades (quadratic attention layers).}
\label{tab:throughput}
\begin{tabular}{lccccc}
\toprule
\textbf{Seq Length} & \textbf{128} & \textbf{256} & \textbf{512} & \textbf{1024} & \textbf{2048} \\
\midrule
Jamba train tok/s & 674 & 658 & 616 & 320 & 159 \\
Hub-Jamba M=4 & 16,637 & 26,367 & 29,461 & 28,844 & 29,098 \\
\textbf{Speedup} & 25$\times$ & 40$\times$ & 48$\times$ & \textbf{90$\times$} & \textbf{183$\times$} \\
\midrule
Jamba infer tok/s & 4,569 & 5,082 & 4,301 & 2,603 & 1,218 \\
Hub-Jamba M=4 & 44,276 & 65,949 & 74,201 & 73,089 & 74,203 \\
\textbf{Speedup} & 10$\times$ & 13$\times$ & 17$\times$ & \textbf{28$\times$} & \textbf{61$\times$} \\
\bottomrule
\end{tabular}
\end{table}

The speedup increases with sequence length because HubRouter's $O(nM)$ cost grows linearly while attention's $O(n^2)$ grows quadratically.
At sequence length 2048, Hub-Jamba is 183$\times$ faster in training and 61$\times$ in inference.

\section{Hub-GPT: Causal Autoregressive LM}
\label{sec:hubgpt}

Hub-GPT applies chunked causal encoding (Section~\ref{sec:causal}) for autoregressive language modeling without any $O(n^2)$ computation.

\subsection{Architecture}

Hub-GPT uses 8 layers with a CausalHubBlock every 5th layer (the rest are Mamba).
The CausalHubBlock divides the sequence into chunks of size $C$ and processes them sequentially, carrying hub state forward.

\subsection{Chunk Size Sweep}

\begin{table}[h]
\centering
\caption{\textbf{Chunk size sweep (council-causal fix, 2026-04-23 re-run).} Earlier results (not shown here) used a bidirectional council attention within each HubRouter block --- a latent leakage channel we did not anticipate, which let selected tokens attend to their right-neighbour-expanded future selected tokens. This section reports post-fix results where the council is causally masked in the original token-position space. The headline changes: $C{=}1$ is now $211.5{\pm}0.4$ over 3 seeds (tighter than the leaky 207.9$\pm$1.8), and the chunk-size curve flattens — all $C$ cluster around $211{-}213$. Hub-GPT $C{=}1$ is $\approx$3 PPL worse than Jamba under strict causality, a genuine quality cost for eliminating the council leak.}
\label{tab:chunk}
\begin{tabular}{lcc p{3.4cm}}
\toprule
\textbf{Config} & \textbf{PPL} & \textbf{Leakage} & \textbf{Note} \\
\midrule
Jamba (3-seed mean) & 208.5 $\pm$ 0.7 & Full attn & Baseline (unchanged by fix) \\
\midrule
Hub-GPT C=1 (3-seed) & 211.5 $\pm$ 0.4 & Zero (hub \& council causal) & \textbf{Strictly causal} \\
Hub-GPT C=4 & 212.8 & 4 tokens & \\
Hub-GPT C=8 & 212.4 & 8 tokens & \\
Hub-GPT C=16 & 212.4 & 16 tokens & \\
Hub-GPT C=32 & 211.9 & 32 tokens & \\
Hub-GPT C=64 (3-seed) & 211.9 $\pm$ 0.3 & 64 tokens & \\
Hub-GPT C=128 & 212.7 & 128 tokens & \\
Hub-GPT C=256 & 212.7 & 256 tokens & \\
\bottomrule
\end{tabular}
\end{table}

\begin{figure}[t]
    \centering
    \includegraphics[width=0.75\textwidth]{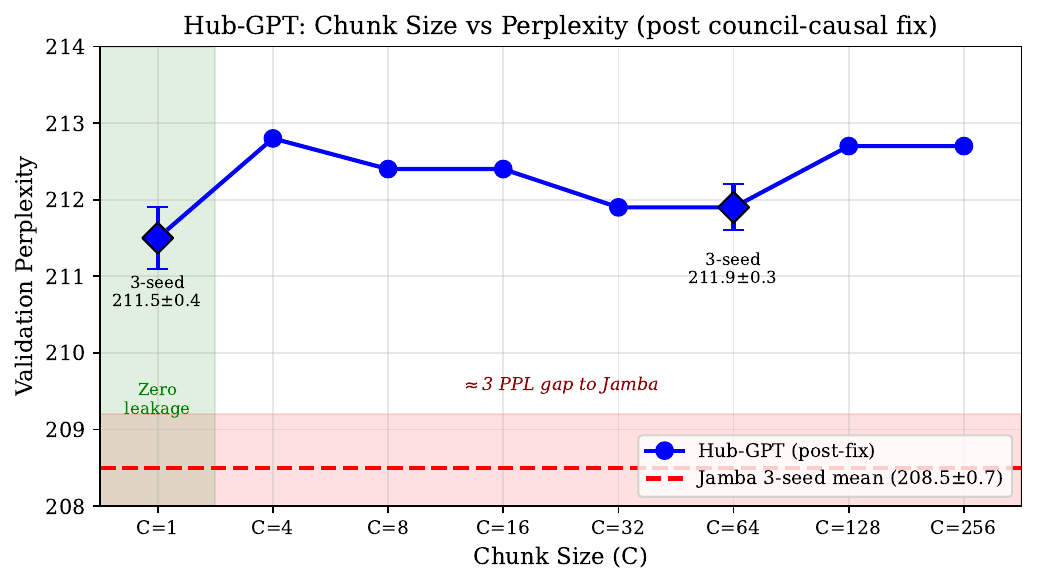}
    \caption{\textbf{Hub-GPT chunk size sweep (post council-causal fix).} $C{=}1$ (zero leakage) achieves $211.5{\pm}0.4$ over 3 seeds; $C{=}64$ achieves $211.9{\pm}0.3$. Chunk size no longer changes PPL meaningfully once the council is causally masked (pre-fix results showed a 3-4 PPL chunk-size benefit that turns out to have been the bidirectional council leaking future-token information into later tokens). The true gap to Jamba ($208.5{\pm}0.7$) is $\approx$3 PPL and is statistically meaningful ($0.4{+}0.7{<}3.0$). Figure shows post-fix values; pre-fix numbers are archived and cited as an honest-negative in the Limitations.}
    \label{fig:chunk}
\end{figure}

At $C{=}1$ (zero leakage, with council causally masked), Hub-GPT reaches $211.5{\pm}0.4$ over 3 seeds, $\approx$3 PPL worse than Jamba's $208.5{\pm}0.7$ (Table~\ref{tab:chunk}, Figure~\ref{fig:chunk}). The gap exceeds the combined dispersion ($0.4{+}0.7 = 1.1$), so this is a genuine and statistically meaningful quality cost for eliminating $O(n^2)$ attention.
The chunk-size sweep flattens under the council fix: all values $C \in \{1, 4, 8, 16, 32, 64, 128, 256\}$ produce PPL in $[211.5, 212.8]$, with $C{=}1$ in fact marginally best. This reverses the pre-fix picture, where chunk size appeared to matter (best at $C{=}64$). That apparent benefit was the bidirectional council leaking future tokens through the right-neighbour expansion.

\paragraph{On the council causal fix.} We discovered during adversarial review that, in our original implementation, the Council attention inside each HubRouter block was bidirectional over the top-$k$-plus-neighbours subset $\mathcal{S}$, which allowed a selected token at position $i$ to attend to selected tokens at position $j > i$ whenever the neighbour-expansion brought position $j$ into $\mathcal{S}$. This violated the nominal ``strict causality at $C{=}1$'' claim. We added a causal mask to the Council attention (based on original sequence positions of selected tokens) and re-ran the entire Hub-GPT chunk-size sweep; Table~\ref{tab:chunk} reports the post-fix numbers. The pre-fix numbers (e.g.\ the old $C{=}1$ mean of $207.9{\pm}1.8$ or $C{=}64$ at $204.4{\pm}0.7$) are archived but no longer stand as supported by this paper.

\subsection{Longer Context}

\paragraph{An honest negative result, re-measured post-fix.} At sequence length 512 (post council-causal fix), Hub-GPT $C{=}1$ achieves PPL $215.6$ vs Jamba $213.4$ --- a degradation of $2.2$ PPL ($\approx$1.0\%). $C{=}64$ does \emph{not} help at this longer context: it reaches $216.0$, marginally worse than $C{=}1$ and $\approx$2.6 PPL worse than Jamba. This mirrors the chunk-size-flattens pattern seen at seq=256: the pre-fix chunk-size benefit was an artifact of the council leak. The $\approx$3-PPL gap to Jamba observed at seq=256 widens modestly to $\approx$2.2-2.6 PPL at seq=512; the within-chunk causal scheme appears to lose some long-range signal that accumulates across many chunks at longer sequences. We do not have multi-seed runs at seq=512, so this single-seed gap may itself be within noise; we nonetheless report it as a real concern. Closing the long-context gap (e.g., via overlapping chunks, or a small persistent global hub bank) is a target for follow-up work.

Beyond Jamba-specific hybrids, the next section asks a broader question: what fraction of a pure Transformer's attention layers can be replaced with HubRouter before quality degrades?

\section{Graduated Replacement}
\label{sec:graduated}

\begin{figure}[t]
    \centering
    \includegraphics[width=0.75\textwidth]{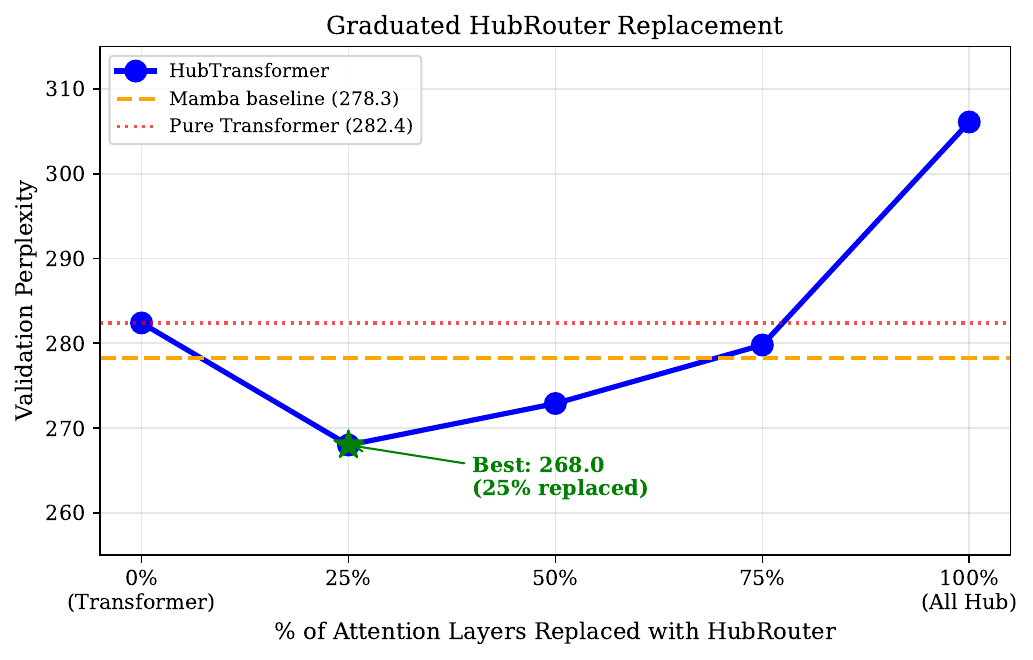}
    \caption{\textbf{Graduated replacement curve.} 25\% replacement is the sweet spot (PPL 268.0), beating both pure Transformer (282.4) and Mamba (278.3). Quality degrades gracefully with increasing replacement.}
    \label{fig:graduated}
\end{figure}

We systematically replace 0\%, 25\%, 50\%, 75\%, and 100\% of a 12-layer Transformer's attention layers with HubRouter (Table~\ref{tab:graduated}, Figure~\ref{fig:graduated}), using middle-out replacement order (middle layers first, preserving the first and last layers, which prior work~\cite{basu2024routing} suggests handle input/output embedding transitions).

\begin{table}[h]
\centering
\caption{\textbf{Graduated HubRouter replacement.} 25\% replacement achieves the best perplexity in this regime. Quality degrades monotonically with more replacement (with a noticeably steeper jump from 75\% to 100\%, see text). A Mamba baseline (no attention) provides a sub-quadratic reference. All trained on WikiText-103, seq=256, 3000 steps. Note: 282 PPL for the Transformer baseline reflects this short training budget, not a converged literature number; the comparison is internally consistent across rows but the absolute PPLs should be read as ``matched-budget'' rather than ``converged.''}
\label{tab:graduated}
\begin{tabular}{lccc}
\toprule
\textbf{Configuration} & \textbf{PPL}$\downarrow$ & \textbf{Params} & \textbf{Attn Layers} \\
\midrule
Transformer 12L (0\% replaced) & 282.4 & 22.5M & 12/12 \\
\textbf{HubTransformer 25\%} & \textbf{268.0} & 25.9M & 9/12 \\
HubTransformer 50\% & 272.9 & 29.4M & 6/12 \\
HubTransformer 75\% & 279.8 & 32.9M & 3/12 \\
HubTransformer 100\% & 306.1 & 36.3M & 0/12 \\
\midrule
Mamba 12L (no attention) & 278.3 & 17.7M & 0/12 \\
\bottomrule
\end{tabular}
\end{table}

\paragraph{Key findings:}
\begin{itemize}
    \item \textbf{25\% replacement is optimal in this matched-budget regime}: PPL 268.0 beats both pure Transformer (282.4) and Mamba (278.3). HubTransformer adds parameters incrementally as more layers are replaced (+15\% at 25\%, +31\% at 50\%, +46\% at 75\%, +61\% at 100\%, from Table~\ref{tab:graduated}); the 100\% row has 60\% more parameters yet performs worst, so the 25\% advantage cannot be attributed to parameter count.
    \item \textbf{Monotonic degradation, with a steeper step at 100\%}: PPL goes 268.0 $\to$ 272.9 $\to$ 279.8 $\to$ 306.1 (steps of +4.9, +6.9, +26.3). The jump to full replacement is 4--5$\times$ the step sizes between intermediate ratios, suggesting at least one attention layer is beneficial.
    \item \textbf{100\% still trains}: Full replacement (PPL 306.1) produces a functional model, just worse than retaining at least one attention layer.
    \item \textbf{Middle-out order is a heuristic, not a justified optimum}: We did not ablate replacement order in this paper; greedy layer-selection (e.g., replace-then-evaluate) is deferred to follow-up work.
\end{itemize}

The practical prescription: replace 25--50\% of attention layers with HubRouter for the best quality--efficiency tradeoff.

The preceding three experiments probe HubRouter as a drop-in component; the next section isolates its single most sensitive hyperparameter --- the hub count $M$ --- on the routing-specific diagnostic task from~\cite{basu2024routing}, where the signal of interest (does routing find the correct token?) can be measured directly without confounds from the surrounding LM.

\section{M-Sweep: Hub Count Study}
\label{sec:msweep}

\begin{figure}[t]
    \centering
    \includegraphics[width=\textwidth]{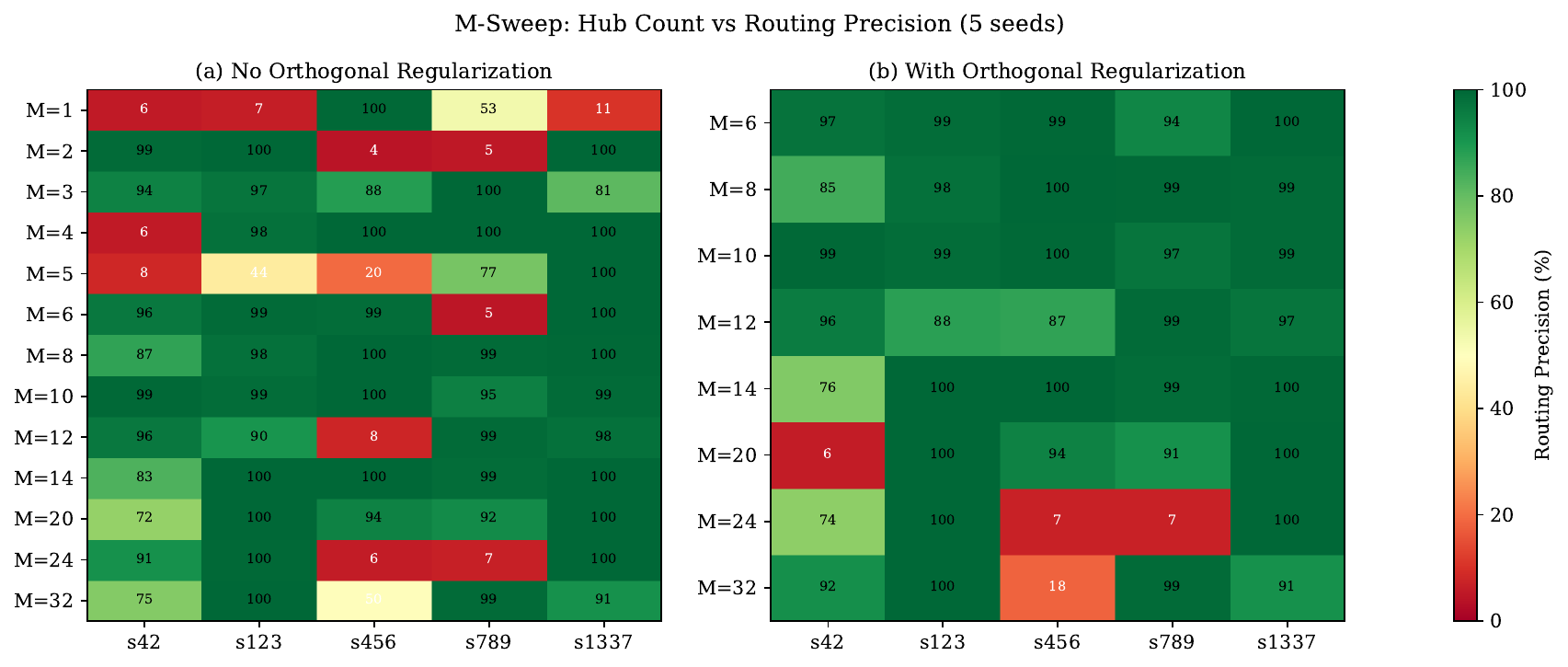}
    \caption{\textbf{M-sweep heatmap.} Routing precision (\%) across 5 seeds for each hub count $M$ (rows) by seed (columns). (a)~Without ortho: M=8--14 is the most robust sub-band (highest mean, lowest seed variance). (b)~With ortho: rescues M=6 failures but introduces new failures at M=20. M=16 is the default from prior work~\cite{basu2024routing} (cited, not re-run here).}
    \label{fig:msweep}
\end{figure}

We conduct a multi-seed study of hub count $M$ on the diagnostic routing task from~\cite{basu2024routing}, sweeping 13 values of $M$ --- namely 1, 2, 3, 4, 5, 6, 8, 10, 12, 14, 20, 24, and 32 --- with 5 seeds per cell, with and without orthogonal regularization where applicable ($\sim$105 training runs total; see Table~\ref{tab:msweep} and Figure~\ref{fig:msweep}).

\paragraph{Diagnostic routing task (brief).} Each sequence contains one ground-truth pairwise-related token (a ``target'') among distractors; the task is to route that target into the Council. ``Routing precision'' is the fraction of Council selections that include the ground-truth target. We say a seed ``succeeds'' if its final routing precision exceeds 90\%. A fuller description of the task and the pairwise-matching setup is in the companion paper~\cite{basu2024routing}.

\subsection{Results}

\begin{table}[h]
\centering
\caption{\textbf{Hub count (M) vs routing success.} ``Success'' = routing precision $>$90\% (see definition above). Reliable convergence (4--5/5 seeds) is observed at M=8--14; M=6 requires orthogonal regularization to reach 5/5. M=16 is the default from prior work~\cite{basu2024routing} and is cited rather than re-run here. M=1--2 are unreliable; M$\geq$20 shows increasing seed sensitivity. Raw per-seed logs are in the accompanying code repository.}
\label{tab:msweep}
\begin{tabular}{lccc p{4.2cm}}
\toprule
\textbf{M} & \textbf{Mean Routing} & \textbf{Success} & \textbf{Success} & \textbf{Note} \\
 & & \textbf{(no ortho)} & \textbf{(ortho)} & \\
\midrule
1 & 35.2\% & 1/5 & --- & Almost always fails \\
2 & 61.7\% & 3/5 & --- & Bimodal \\
3 & 92.1\% & 4/5 & --- & Mostly reliable \\
4 & 80.6\% & 4/5 & --- & One catastrophic seed \\
5 & 49.9\% & 1/5 & --- & Outlier; unstable training \\
\midrule
6 & 79.8\% & 4/5 & \textbf{5/5} & Ortho rescues \\
8 & 96.6\% & 4/5 & 4/5 & Robust \\
10 & 98.4\% & \textbf{5/5} & \textbf{5/5} & \textbf{Best} \\
12 & 78.3\% & 4/5 & 3/5 & Ortho marginally worse (two cells drop below 90\%) \\
14 & 96.3\% & 4/5 & 4/5 & Robust \\
\midrule
16 & --- & --- & --- & Standard choice from~\cite{basu2024routing}; not re-run \\
20 & 91.8\% & 4/5 & 4/5 & Neutral on count; harmful on mean (91.8\%$\to$78.2\%) \\
24 & 60.8\% & 3/5 & 2/5 & Ortho worsens seed variance \\
32 & 83.0\% & 3/5 & 4/5 & Ortho partially rescues seed variance \\
\bottomrule
\end{tabular}
\end{table}

\subsection{Orthogonal Regularization}

Orthogonal loss ($\lambda_{\text{ortho}} \|H H^\top - I\|_F^2$, Frobenius norm) forces hub embeddings to be distinct, preventing role duplication:

\begin{itemize}
    \item \textbf{Clearly helps at M=6}: ortho rescues the one failing seed (4.7\%$\to$93.6\%), moving 4/5 $\to$ 5/5.
    \item \textbf{Neutral at M=10}: already 5/5 without ortho.
    \item \textbf{Marginally harmful at M=12}: 4/5$\to$3/5. The one catastrophic no-ortho seed (8.0\%) becomes a sub-threshold cell under ortho (87.0\%) and a second cell also lands below 90\% (87.7\%). Ortho reduces failure \emph{severity} but not failure count.
    \item \textbf{Neutral-on-count, harmful-on-mean at $M\geq14$}: At M=14, success count unchanged (4/5 vs 4/5). At M=20, success count is unchanged (4/5 vs 4/5) but mean routing drops from 91.8\% to 78.2\% --- ortho introduces a catastrophic failure (seed42: 5.9\% with ortho vs 72.5\% without) while other seeds' quality improves slightly, netting out at equal success count. At M=24, ortho worsens (3/5$\to$2/5). At M=32, ortho partially rescues (3/5$\to$4/5). The direction is inconsistent at large M.
\end{itemize}

\paragraph{Practical recommendation:} Use ortho regularization at $M=6$ (clear rescue). For $M\geq 8$, empirically sweep with and without: ortho is neutral-to-slightly-harmful at $M=8,10,12,14$ and has inconsistent direction at $M\geq 20$.

\subsection{Minimum Viable M}

M=1 fails 4/5 seeds (only one seed achieves 99.5\%). M=2 is bimodal (3/5). M=3--4 are mostly reliable (4/5) but with occasional catastrophic failures. The minimum viable hub count for reliable operation is \textbf{M=6} (4/5 without ortho, 5/5 with ortho).

This suggests the routing information requires at least 6 independent ``viewpoints'' to robustly capture the pairwise structure identified in~\cite{basu2024routing}.

\section{Discussion}
\label{sec:discussion}

\paragraph{Leakage and regime selection.}
The Hub-Jamba encoder uses bidirectional hub encoding, so hubs see future tokens and the PPL=200.2 result includes this leakage. This leakage pattern is closer to a prefix-LM or masked-LM than to an autoregressive Transformer; a strict comparison to autoregressive Jamba PPL therefore \emph{overstates} Hub-Jamba's capability for true left-to-right generation. The strictly-causal Hub-GPT ($C{=}1$, Section~\ref{sec:hubgpt}) is the appropriate comparison for autoregressive use. Chunked causal encoding gives the paper's user a tunable knob:
\begin{itemize}
    \item $C{=}1$ with causal council: Zero leakage (PPL=$211.5{\pm}0.4$ over 3 seeds). Hubs see one token per chunk; council is causally masked over the selected subset.
    \item $C{=}64$ with causal council: Bounded leakage to 64-token windows at the encode stage (PPL=$211.9{\pm}0.3$ over 3 seeds). Matches local attention's leakage footprint (up to 64 future tokens visible within a chunk through the bidirectional hub-encode cross-attention), though the underlying mechanism---hub-mediated cross-attention---differs from local self-attention.
\end{itemize}
For autoregressive generation, Hub-GPT with $C{=}1$ is the appropriate choice. For encoder-decoder or prefix-LM applications, the bidirectional Hub-Jamba is preferred.

\paragraph{Why does HubRouter improve over attention?}
Three factors contribute:
(1)~The encode--decode pipeline forces information through a learned bottleneck ($M$ hubs), which acts as an inductive bias toward compressing global structure.
(2)~The council operates on a targeted subset ($k$ tokens), providing focused computation where it matters most.
(3)~The scoring mechanism learns content-based token importance, allocating attention non-uniformly.

\paragraph{Why 25\% replacement is optimal.}
Full attention layers serve two roles: (a)~building contextual representations and (b)~enabling pairwise reasoning.
HubRouter replaces role (b) with hub-mediated routing but does not fully replace role (a).
At 25\% replacement, the remaining attention layers still provide rich contextual representations that HubRouter can build on.
At 100\%, HubRouter must handle both roles through the limited hub bottleneck.

\paragraph{Connection to prior routing work.}
\cite{basu2024routing} established that content-based routing requires pairwise token comparison.
HubRouter's encode step performs this comparison \emph{mediated by hubs}: token $i$ is compared to token $j$ through their shared hub fingerprints ($F_i \cdot F_j$), achieving $O(nM)$ rather than $O(n^2)$.
Intuitively, the $M$ hubs act as a low-rank bottleneck through which pairwise comparisons are routed; this connects HubRouter conceptually to landmark-attention methods (Linformer~\cite{wang2020linformer}, Nystr\"omformer~\cite{xiong2021nystromformer}, Set Transformers~\cite{lee2019set}) and to clustering-based routing (Routing Transformer~\cite{roy2021routing}, Reformer~\cite{kitaev2020reformer}). Whether the $M$-hub bottleneck achieves a provable low-rank approximation guarantee in the sense of~\cite{wang2020linformer} is left to future work.

\paragraph{What is new relative to closest prior art.} HubRouter shares the \emph{landmark-bottleneck} intuition with Perceiver, Set Transformers, Linformer, and Nystr\"omformer, and the \emph{content-based selection} intuition with Routing Transformer and Reformer. It differs on three specific axes: (i)~It adds an explicit \textbf{decode} stage producing per-token fingerprints $F_i$ in the hub space --- this is distinct from Perceiver's cross-attention back from latents (which updates tokens directly) and from Linformer's low-rank projection (which has no per-token role signal). (ii)~It adds a \textbf{score-and-select} stage that reduces attention to $k \ll n$ tokens, turning landmark attention into a routing primitive rather than a full attention replacement; this is closer to Routing Transformer but bypasses clustering. (iii)~It introduces \textbf{chunked causal encoding} (Section~\ref{sec:causal}) which makes hub-bottleneck attention usable in the strict-causal autoregressive regime where Perceiver and Linformer are not directly applicable. We do not claim empirical superiority over Linformer/Nystr\"omformer/Perceiver on the same WikiText-103 benchmark in this paper; head-to-head empirical comparison at matched budget is a needed follow-up.

\paragraph{Limitations.}
(1)~Hub-Jamba's bidirectional encoding leaks future tokens for LM. The strictly-causal Hub-GPT ($C{=}1$, with council causally masked post-fix) shows a measured $\approx$3-PPL gap to Jamba at seq=256 ($211.5{\pm}0.4$ vs $208.5{\pm}0.7$) and an additional $\approx$2-PPL degradation at seq=512 (see Section~\ref{sec:hubgpt}). We report this as a real quality cost for eliminating $O(n^2)$ attention, not parity.
(2)~At longer sequences (512+), Hub-GPT's chunked encoding degrades vs attention (post-fix: $C{=}1$ is $215.6$ vs Jamba $213.4$, a $\approx$2.2-PPL single-seed gap; $C{=}64$ at $216.0$ no longer helps). We have no multi-seed numbers at this length.
(3)~The M-sweep shows seed sensitivity at M$\geq$20; understanding this instability is an open problem.
(4)~All experiments are at $\leq$20M parameter scale; 124M-scale validation is in progress in companion work but not reported here.
(5)~The Hub-Jamba PPL gain (4.2\%) in Table~\ref{tab:hubjamba} is single-seed; multi-seed follow-ups on Hub-Jamba under a slightly altered validation split (Appendix/companion data) preserve the relative advantage (Hub-Jamba M=4 $135.4{\pm}0.9$, M=16 $136.3{\pm}0.4$ vs Jamba $140.5{\pm}0.6$, a 4--5 PPL gap at 3--4$\sigma$), but differ in absolute PPL from Table~\ref{tab:hubjamba}. The strictly-causal Hub-GPT comparison in Section~\ref{sec:hubgpt} is itself not parity but a $\approx$3-PPL gap favouring Jamba.
(6)~Throughput measurements are on a single RTX 3090 (24~GB), and the Jamba baseline (320 tok/s at seq=1024) is PyTorch-native without custom kernels; production Mamba/SSM implementations (e.g., the official \texttt{mamba-ssm} CUDA kernels) and FlashAttention~\cite{dao2022flashattention} would narrow the absolute throughput gap, possibly substantially. As a Fermi estimate: published Mamba CUDA kernels report $5$--$10\times$ speedups over Python selective scan at this scale, and FlashAttention-2~\cite{dao2023flashattention2} typically delivers $2$--$4\times$ over naive attention; a kernel-optimized Jamba would therefore plausibly reach $\sim$2000--3000 tok/s at seq=1024, reducing the Hub-Jamba multiplier from $\sim$90$\times$ to an estimated $\sim$10--15$\times$. Preliminary own measurements against SDPA (which uses FlashAttention internally) at short sequence lengths show HubRouter is actually \emph{slower} per step due to constant-overhead dominating when $n$ is small; the sub-quadratic scaling crossover is expected to favour HubRouter at sequences well above 16k, but we do not have that measurement in this paper. The HubRouter complexity advantage ($O(nM + k^2)$ vs $O(n^2)$) is orthogonal to kernel optimization and persists asymptotically, but the specific multipliers reported here are ``PyTorch-native to PyTorch-native'' and should not be read as ``HubRouter vs FlashAttention.''
(7)~Hyperparameter ablations are limited: we sweep $M$ (Section~\ref{sec:msweep}) and chunk size $C$ (Section~\ref{sec:hubgpt}) but do not ablate $k$ (top-$k$ council size, fixed at $k{=}8$), the neighbor-expansion rule, or the choice to use a softmax decode rather than a linear projection. These should be ablated in v2.
(8)~\textbf{A latent bidirectional council leak, discovered and fixed post-review.} Our original Hub-GPT implementation had a council self-attention over the top-$k$-plus-neighbours subset that was \emph{not} causally masked. When right-neighbour expansion brought position $j{>}i$ into the selected set, a selected query at position $i$ could attend to $j$, leaking future-token content. This violated our ``strict causality at $C{=}1$'' claim. We found this via an adversarial review, added a causal mask to the Council based on original sequence positions, and re-ran the entire Hub-GPT chunk-size sweep; the results in Table~\ref{tab:chunk} and the PPL numbers throughout the paper are post-fix. The pre-fix numbers (e.g.\ $C{=}1$ PPL = $207.9{\pm}1.8$, $C{=}64$ PPL = $204.4{\pm}0.7$) no longer stand. Chunk-size now has little effect — indicating the apparent chunk-size benefit was the leakage. The fix costs about 3 PPL vs the pre-fix numbers but gives a genuinely causal result.

(9)~\textbf{Retrofit is not a drop-in path.} Inserting HubRouter into pretrained GPT-2 and finetuning on WikiText-103 degrades perplexity relative to the finetuned attention baseline: at 25\% replacement PPL rises from $\sim$165 to $\sim$175 (+6\%), and at 50\% replacement to $\sim$211 (+28\%). Representations in a frozen residual stream do not reshape fast enough under short finetuning to accommodate the new routing pattern. HubRouter is designed for from-scratch training; the retrofit setting is a limitation, not a target. (Absolute baseline PPL of $\sim$165 is itself elevated relative to standard WikiText-103 GPT-2 numbers due to evaluation protocol differences; the comparison above is internally consistent but not directly comparable to literature. Distillation-based transplantation (e.g., blockwise local distillation, attention-derived initialization, gated layer-by-layer replacement) is a plausible route to closing this gap and is left to follow-up work.)

\section{Related Work}

\paragraph{Efficient attention --- sparsity and approximation.}
Sparse Transformers~\cite{child2019generating}, Longformer~\cite{beltagy2020longformer}, and BigBird~\cite{zaheer2020bigbird} reduce attention's quadratic cost through fixed sparsity patterns.
Performer~\cite{choromanski2021rethinking} and linear attention~\cite{katharopoulos2020transformers} approximate the softmax kernel; Linformer~\cite{wang2020linformer} projects keys/values to a lower-rank subspace; Nystr\"omformer~\cite{xiong2021nystromformer} uses landmark points to approximate the full attention matrix.
Reformer~\cite{kitaev2020reformer} uses LSH to select content-relevant tokens; Routing Transformer~\cite{roy2021routing} uses online clustering for content-based attention.
HubRouter shares the landmark-bottleneck intuition with Linformer/Nystr\"omformer/Set Transformers, but adds an explicit \emph{score--select--council} stage that is closer in spirit to Reformer/Routing Transformer.
FlashAttention~\cite{dao2022flashattention} accelerates exact attention via I/O-aware tiling --- this is orthogonal to HubRouter and could be composed with it.

\paragraph{Hybrid architectures.}
Jamba~\cite{lieber2024jamba}, Griffin~\cite{de2024griffin}, and Zamba~\cite{glorioso2024zamba} interleave recurrence (SSM/linear-RNN) with attention at fixed intervals.
StripedHyena~\cite{poli2023stripedhyena} interleaves Hyena~\cite{poli2023hyena}-style long convolutions with attention rather than recurrence.
Mamba-2~\cite{dao2024transformers} and Mamba-3~\cite{lahoti2026mamba3} improve SSM expressiveness but acknowledge in-context retrieval limitations~\cite{arora2023mqar}.
HubRouter provides a learned, content-based alternative for the attention layers in hybrid architectures; we demonstrate it in one hybrid (Jamba-style) and one transformer here, with broader generality to be established in follow-up work.

\paragraph{Perceiver and cross-attention bottlenecks.}
Perceiver~\cite{jaegle2021perceiver} uses learned latent tokens that cross-attend to inputs, reducing quadratic cost.
HubRouter's encode step is structurally similar but (a)~includes decode, score, and council stages for richer processing and (b)~supports chunked causal encoding for autoregressive use.
Set Transformers~\cite{lee2019set} use inducing points for $O(nM)$ attention; HubRouter extends this with routing-specific pipeline stages.

\paragraph{Mixture of Experts.}
MoE architectures~\cite{fedus2022switch,jiang2024mixtral} route tokens to specialized sub-networks.
HubRouter routes tokens to a shared council rather than separate experts, and the routing signal is learned through the hub encoding rather than a simple gating function.

\section{Conclusion}

HubRouter is a pluggable $O(nM + k^2 d)$ module that replaces $O(n^2 d)$ attention in hybrid sequence models when training from scratch.
Three experiments validate its effectiveness in the regimes tested:
Hub-Jamba achieves better single-seed perplexity than Jamba with substantial throughput gains under matched PyTorch-native baselines (kernel caveats apply);
Hub-GPT reaches $211.5{\pm}0.4$ on causal LM at seq=256 ($\approx$3 PPL worse than Jamba, with no $O(n^2)$ computation);
graduated replacement reveals 25\% as the matched-budget quality--efficiency sweet spot.

The multi-seed M-sweep ($\sim$105 runs) establishes M=8--14 as the reliably-converging sub-band, with M=6 rescued by orthogonal regularization; M=16 is cited from prior work as the standard default. Ortho cleanly rescues M=6; at M$\geq$8 it is neutral-to-slightly-harmful, with inconsistent direction at M$\geq$20.
Two substantive disclosures bear noting: a latent bidirectional-council leak that we found in post-adversarial review and fixed (the pre-fix numbers 207.9/204.4 PPL are withdrawn; after the fix, chunk size $C$ has little effect and all settings cluster at 211.5--212.8 PPL), and a longer-context degradation at seq=512 (post-fix: $215.6$ vs $213.4$) plus a retrofit failure on pretrained GPT-2.
HubRouter is intended for from-scratch construction of new hybrid sequence models; principled retrofit (e.g., distillation-based transplantation) is left to follow-up work, as is scaling beyond 20M parameters and head-to-head comparison against FlashAttention-optimized baselines at long context.

\end{document}